\documentclass[conference]{IEEEtran}

\usepackage{cite}
\usepackage{amsmath,amssymb,amsfonts}
\usepackage{graphicx}
\usepackage{textcomp}
\usepackage{xcolor}
\usepackage{booktabs}
\usepackage{multirow}
\usepackage{CJKutf8}
\usepackage{hyperref}
\usepackage{tcolorbox}
\tcbuselibrary{skins}
\usetikzlibrary{shapes.geometric, arrows.meta, positioning, fit, backgrounds}

\hypersetup{
    colorlinks=false,
    hidelinks
}

\begin{document}

\title{Emotion-Aware Image Generation from Korean Diary Text via LLM-based Prompt Translation and LoRA Fine-Tuning\\
{\large A Children's Drawing Style Pipeline for Korean Emotional Diaries}}

\author{
\IEEEauthorblockN{Jihun Cho}
\IEEEauthorblockA{Dept. of Computer Science\\
Pai Chai University\\
Daejeon, Korea\\
2161083@pcu.ac.kr}
\and
\IEEEauthorblockN{Soo-Yeon Jeong}
\IEEEauthorblockA{Smart ICT Convergence Human Resources\\
Development Center\\
Pai Chai University\\
Daejeon, Korea\\
sy.jeong@pcu.ac.kr}
\and
\IEEEauthorblockN{Sun-Young Ihm}
\IEEEauthorblockA{Dept. of Computer Science\\
Pai Chai University\\
Daejeon, Korea\\
sunnyihm@pcu.ac.kr\\
\textit{Corresponding Author}}
}

\maketitle

\begin{abstract}
T2I models cannot effectively capture sentiment from various types of text, including diaries, as they primarily focus on visual object-related patterns rather than contextual emotional understanding. This paper proposes an emotion-aware text-to-image pipeline that generates children's hand drawing style images from short Korean diary entries. The proposed pipeline employs Qwen3-8B for recognising implicit sentiment from short diaries, and Stable Diffusion 3.5 Medium fine-tuned with LoRA on children's drawing images with emotion-based trigger words for image generation. Additionally, this paper presents experiments examining the effect of emotion trigger words on generated images and discusses the limitations of CLIP Score as an evaluation metric for emotion-aware image generation.
\end{abstract}

\begin{IEEEkeywords}
Text-to-Image Generation; LoRA Fine-Tuning; Emotion-Aware; Korean Diary; Children's Drawing Style; Stable Diffusion
\end{IEEEkeywords}

% -------------------------------------------------------
\section{Introduction}

T2I models struggle to represent sentiment in diverse forms of text because emotional concepts are more abstract than concrete visual concepts, which existing diffusion models handle more effectively~\cite{emogen}. Among these, diary entries are particularly challenging, as they inherently require sentiment understanding to accurately capture the emotional tone intended by the writer. Diary text frequently contains implicit emotional context expressed through short and subjective descriptions, making it difficult for existing T2I models to recognise and represent the intended emotional nuance from the text alone. Children's drawings are often used as expressive visual representations that communicate feelings and intentions rather than purely realistic depictions~\cite{childrendrawings}, making them a suitable visual medium for representing the emotional content of diary entries.

This paper proposes a pipeline for generating children's drawing style images from Korean diary entries, which is capable of recognising and reflecting implicit sentiment from the text. The proposed pipeline mainly consists of two stages: sentiment recognition using the zero-shot LLM Qwen3-8B, and image generation using Stable Diffusion 3.5 Medium trained on emotion-inclusive children's drawing datasets with emotion-based trigger words. Through experiments, it is demonstrated that emotion trigger words significantly affect the visual emotional expression of generated images, and that CLIP Score has limitations in evaluating style transfer quality. Furthermore, while the current pipeline addresses the language barrier through LLM-based translation, direct Korean-language input processing, Korean text rendering in generated images, and the expansion of emotion categories beyond happiness and sadness remain topics for future research.

% -------------------------------------------------------
\section{Related Work}

\subsection{Text-to-Image Generation}

Text-to-image (T2I) generation has rapidly advanced with the development of diffusion-based models, leading to many commercialised systems such as DALL$\cdot$E 3~\cite{dalle3} and Midjourney~\cite{midjourney}. Diffusion-based T2I models have shown remarkable progress in generating high-quality images from text, with recent models such as Stable Diffusion 3.5~\cite{sd35} demonstrating improved text understanding and image quality. However, these models primarily focus on visual object-level patterns, rather than abstract emotional concepts. This has motivated research into sentiment-aware T2I generation.

\subsection{LoRA Fine-Tuning}

LoRA fine-tuning~\cite{lora} is a parameter-efficient approach that enables model adaptation with a limited amount of data. Unlike conventional fine-tuning, which requires training the entire model on large datasets, LoRA trains a small set of additional weights, allowing the model to learn a particular style or feature while keeping the original parameters frozen. In the T2I field, LoRA has been widely used to condition models on specific visual concepts or styles. Its parameter-efficient design makes LoRA particularly suitable for learning specialised artistic styles from limited training data.

\subsection{Sentiment Analysis}

Sentiment analysis has evolved through several stages: dictionary-based methods that rely on predefined word scores, machine learning approaches that learn patterns from labelled data, and transformer-based models such as BERT that capture contextual understanding. More recently, LLMs have demonstrated the ability to perform sentiment analysis in a zero-shot manner, recognising implicit emotional expressions without requiring task-specific training data~\cite{llmsentiment}. This makes them a viable approach for analysing diverse and informal text such as Korean diary entries.

\subsection{Emotion in Image Generation}

Several studies have explored emotional image generation by incorporating emotional information into text-to-image generation models. Notably, EmoGen~\cite{emogen} generates images based on emotion categories such as happiness, sadness, fear, and disgust, introducing a new task termed ``Emotional Image Content Generation (EICG)''. However, these approaches primarily rely on explicit emotional labels and English-language text, limiting their applicability to non-English and emotionally nuanced texts such as diary entries.

% -------------------------------------------------------
\section{Method}

\subsection{Pipeline Overview}

\begin{figure}[htbp]
\centering
\begin{tikzpicture}[
    node distance=0.45cm,
    box/.style={rectangle, rounded corners=6pt, draw=black, minimum width=4.8cm,
                minimum height=1.1cm, align=center, font=\small},
    subtext/.style={font=\scriptsize},
    arrow/.style={-{Stealth[length=7pt]}, thick}
]
\node[box, fill=blue!20] (diary)
    {\textbf{Korean Diary}\\[1pt]
     {\scriptsize 2--5 sentences, $\sim$100 chars}};
\node[box, fill=green!20, below=of diary] (llm)
    {\textbf{Qwen3-8B}\\[1pt]
     {\scriptsize Emotion detection $\rightarrow$ English prompt}};
\node[box, fill=orange!20, below=of llm] (sd)
    {\textbf{SD 3.5 Medium}\\[1pt]
     {\scriptsize + LoRA weights, 512$\times$512, 28 steps}};
\node[box, fill=teal!20, below=of sd] (image)
    {\textbf{Generated Image}\\[1pt]
     {\scriptsize Children's drawing style}};
\draw[arrow] (diary) -- (llm);
\draw[arrow] (llm) -- (sd);
\draw[arrow] (sd) -- (image);
\end{tikzpicture}
\caption{Overview of the proposed Diary2Image pipeline.}
\label{fig:pipeline}
\end{figure}
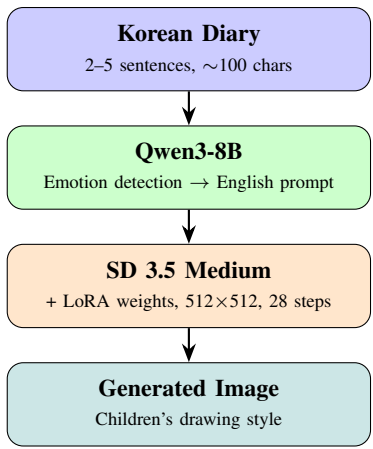

The proposed pipeline consists of four stages, as illustrated in Fig.~\ref{fig:pipeline}. A short Korean diary entry is first passed to Qwen3-8B, which detects the emotional tone and converts it into a detailed English prompt with an emotion-based trigger word. The generated prompt is then fed into Stable Diffusion 3.5 Medium fine-tuned with LoRA to generate a children's hand drawing style image.

\subsection{Prompt Generation}

In the prompt generation stage, Qwen3-8B, a large language model (LLM) developed by Alibaba Cloud, is utilised. It converts a short Korean diary entry into a detailed English prompt for image generation, focusing on scene, mood, atmosphere, lighting, colours, and objects. Additionally, the model generates an emotion trigger word, either \textit{happiness} or \textit{sadness}, based on the sentiment detected from the diary. The emotion trigger is prepended to the prompt, forming inputs such as ``hand drawing happiness style, a quiet afternoon with rain tapping on the window...''. Unlike dictionary-based and transformer-based models that require task-specific fine-tuning on sentiment analysis datasets, Qwen3-8B recognises emotion in a zero-shot manner. An example of the prompt generation process is illustrated in Fig.~\ref{fig:promptgen}.

\begin{figure}[htbp]
\centering
\begin{tcolorbox}[
    colback=white,
    colframe=gray!60,
    colbacktitle=gray!40,
    title={\small Fig. 2. Prompt Generation Example},
    fonttitle=\small,
    boxrule=0.5pt,
    top=4pt, bottom=4pt, left=4pt, right=4pt
]
\small
\textbf{Input (Korean):}
\begin{CJK}{UTF8}{mj}빗소리가 창문을 두드렸다. 고양이는 내 옆에 딱 붙어서 잠들었고, 나도 그냥 같이 누워버렸다. 아무 말 없이도 충분한 오후였다.\end{CJK}

\textbf{Input (English):} The rain tapped on the window. My cat curled up beside me and fell asleep, and I just lay down with it. It was a sufficient afternoon without a word.

\smallskip
\textbf{Output Prompt:} \textit{hand drawing happiness style, a peaceful afternoon scene, rain gently tapping on the window, a cat curled up next to me, both lying on the floor in a cozy living room, warm and soft lighting, pastel colors, a calm and comforting atmosphere, the sound of rain creating a soothing background, a sense of quiet contentment, soft shadows, fluffy clouds outside the window, a warm yellow light from the lamp, a blanket draped over the back, a feeling of safety and togetherness}
\end{tcolorbox}
\caption{Prompt Generation Example.}
\label{fig:promptgen}
\end{figure}

\subsection{Image Generation}

In the image generation stage, Stable Diffusion 3.5 Medium was selected due to its lower computational requirements compared to larger variants while maintaining competitive image quality. The English prompt generated in the previous stage is passed to the model, which produces a 512$\times$512 image using 28 inference steps. A guidance scale of 7.0 is applied to balance the trade-off between prompt adherence and image diversity. The pre-trained LoRA weights are loaded via the diffusers library, conditioning the model on the children's hand drawing style.

\subsection{LoRA Training}

The KIDO dataset~\cite{kido} was selected for LoRA fine-tuning since it contains children's hand drawing images explicitly associated with two emotional categories, happiness and sadness. A total of 9,228 images were used in this study, consisting of 4,614 happiness images and 4,614 sadness images. The dataset provides emotionally expressive visual characteristics that are suitable for training a sentiment-aware image generation model. LoRA fine-tuning was conducted for 10,000 steps with a learning rate of 4e-4 using the diffusers training pipeline. Each image in the dataset is paired with a text caption containing an emotion-based trigger word, which conditions the model to associate specific visual styles with emotional expressions. The details of the caption strategies are described in Section~\ref{sec:experiment}.

% -------------------------------------------------------
\section{Experiment}
\label{sec:experiment}

\subsection{Experiment Setup}

Experiments were conducted on 30 Korean diary entries, including happiness, sadness, and neutral entries. Generated images are evaluated using two metrics: CLIP Score and style transfer rate. CLIP Score is computed using LongCLIP~\cite{longclip}, which supports up to 248 tokens, making it suitable for evaluating the longer and richer prompts generated from diary text. Style transfer rate refers to the proportion of generated images that visually exhibit children's hand drawing characteristics, assessed through manual inspection. All experiments were conducted using the HuggingFace diffusers library. LoRA fine-tuning was performed on an NVIDIA RTX 5090 via RunPod cloud, while image generation was performed locally on an NVIDIA RTX 3090.

\subsection{Effect of Learning Rate}

Two different learning rates were examined in this experiment: 1e-4, which is commonly used for LoRA fine-tuning, and 4e-4, which is four times higher, applied due to the limited size of the training dataset. Overall, the average CLIP Score of 1e-4 (25.64) was higher than that of 4e-4 (25.00). However, this did not indicate that 1e-4 produced higher quality images, as only an average of 69\% of images generated with 1e-4 exhibited the children's drawing style, compared to 81\% for 4e-4, which successfully applied the style. Furthermore, even when the style was applied, it was often incomplete. This suggested that 1e-4 was insufficient to learn the abstract characteristics of children's drawings, resulting in more realistic images that tend to achieve higher CLIP Scores, rather than reflecting genuine style transfer quality. Therefore, 4e-4 was selected for all subsequent experiments. The results are summarised in Table~\ref{tab:lr}.

\begin{table}[htbp]
\caption{Effect of Learning Rate}
\label{tab:lr}
\centering
\begin{tabular}{llccc}
\toprule
\textbf{Caption Strategy} & \textbf{LR} & \textbf{CLIP} & \textbf{Std} & \textbf{Style Rate} \\
\midrule
\multirow{2}{*}{Trigger Only}   & 4e-4 & 25.10 & 0.89 & 77\% \\
                                 & 1e-4 & 25.78 & 0.88 & 50\% \\
\midrule
\multirow{2}{*}{Emotion Style}  & 4e-4 & 24.66 & 1.39 & 97\% \\
                                 & 1e-4 & 25.56 & 1.21 & 90\% \\
\midrule
\multirow{2}{*}{Emotion + Text} & 4e-4 & 25.25 & 1.10 & 70\% \\
                                 & 1e-4 & 25.58 & 1.13 & 67\% \\
\bottomrule
\end{tabular}
\end{table}

\subsection{LoRA Caption Strategy and Trigger Word Comparison}

Three different LoRA caption strategies were compared, each trained with a learning rate of 4e-4 for 10,000 steps. In the Trigger Only strategy, all images were captioned with a single fixed trigger word \textit{kido style}, regardless of their emotional label. In the Emotion Style strategy, captions were assigned based on the emotional category of each image, using either \textit{kido happiness style} or \textit{kido sadness style}. In the Emotion + Text strategy, the emotional trigger was further combined with the child's original emotion description text, forming captions such as ``kido sadness style, being alone always makes me feel sad''. During inference, the trigger word was changed from \textit{kido} to \textit{hand drawing}, resulting in all experiments achieving a style transfer rate of 100\%. For inference, the Trigger Only strategy used \textit{hand drawing style}, while the Emotion Style and Emotion + Text strategies used \textit{hand drawing happiness/sadness style} based on the detected sentiment. The results are summarised in Table~\ref{tab:caption}.

\begin{table}[htbp]
\caption{LoRA Caption Strategy and Trigger Word Comparison}
\label{tab:caption}
\centering
\begin{tabular}{lccc}
\toprule
\textbf{Caption Strategy} & \textbf{CLIP} & \textbf{Std} & \textbf{Style Rate} \\
\midrule
Trigger Only   & 26.73 & 1.04 & 100\% \\
Emotion Style  & 25.95 & 0.95 & 100\% \\
Emotion + Text & 26.15 & 0.84 & 100\% \\
\bottomrule
\end{tabular}
\end{table}

\begin{figure}[htbp]
\centering

\begin{tcolorbox}[colback=white, colframe=gray!60, boxrule=0.5pt, top=3pt, bottom=3pt, left=3pt, right=3pt]
\small
\textbf{Diary 1 (Sadness):} \begin{CJK}{UTF8}{mj}할머니 댁에 가는 길에 창밖을 한참 봤다. 예전엔 자주 갔는데 이제는 명절에만 간다. 그 사실이 이상하게 마음에 걸렸다.\end{CJK}\\
\textit{On the way to grandma's house, I stared out the window for a long time. I used to go often, but now only on holidays. That fact weighed on me strangely.}
\end{tcolorbox}

\vspace{2pt}
\begin{tabular}{ccc}
\includegraphics[width=0.3\columnwidth]{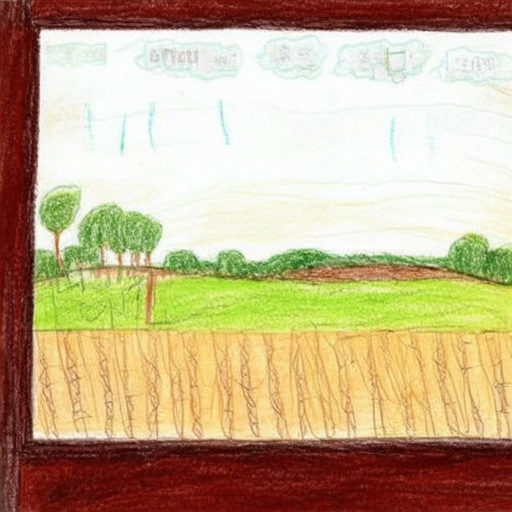} &
\includegraphics[width=0.3\columnwidth]{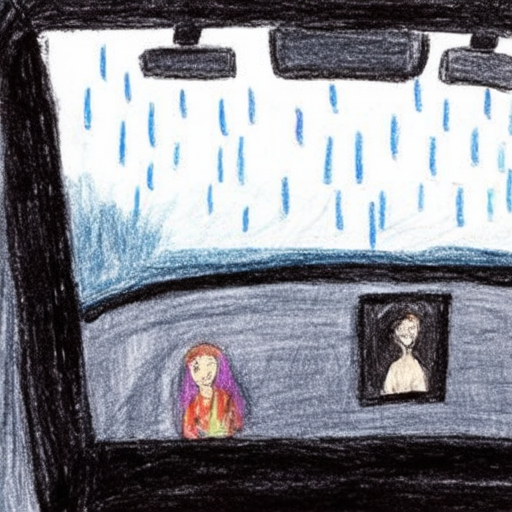} &
\includegraphics[width=0.3\columnwidth]{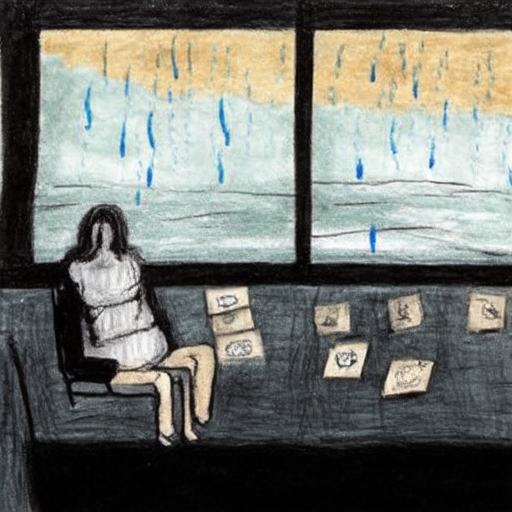} \\
{\small Trigger Only} & {\small Emotion Style} & {\small Emotion + Text}
\end{tabular}

\vspace{4pt}
\begin{tcolorbox}[colback=white, colframe=gray!60, boxrule=0.5pt, top=3pt, bottom=3pt, left=3pt, right=3pt]
\small
\textbf{Diary 2 (Sadness):} \begin{CJK}{UTF8}{mj}오늘 길에서 예전에 자주 가던 가게가 문을 닫은 걸 봤다. 그냥 지나쳤는데 자꾸 생각났다. 이런 것들이 쌓이는 것 같다.\end{CJK}\\
\textit{I saw that a shop I used to frequent had closed. I just walked past, but it kept coming back to me. These things seem to accumulate.}
\end{tcolorbox}

\vspace{2pt}
\begin{tabular}{ccc}
\includegraphics[width=0.3\columnwidth]{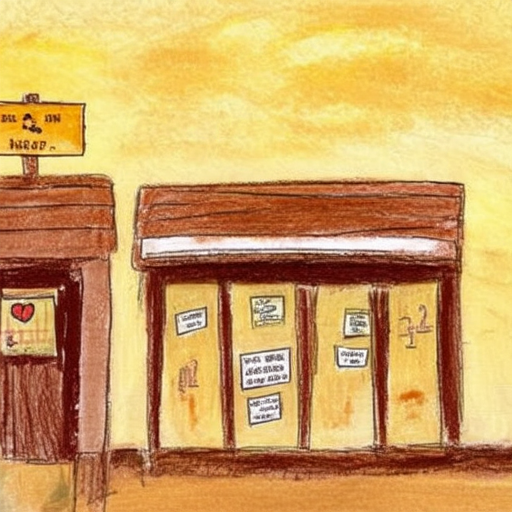} &
\includegraphics[width=0.3\columnwidth]{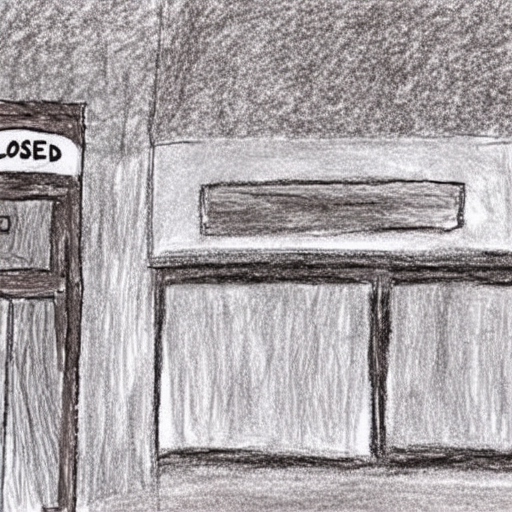} &
\includegraphics[width=0.3\columnwidth]{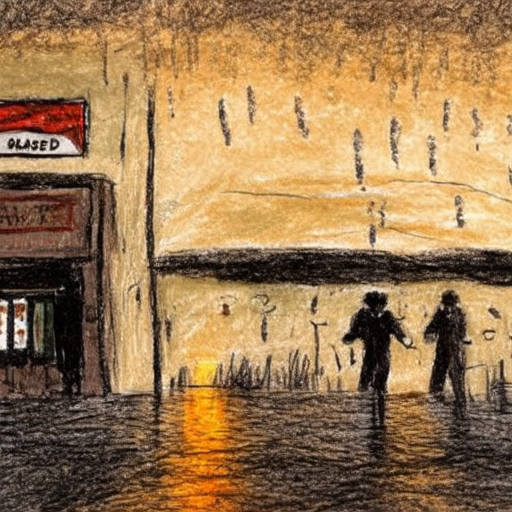} \\
{\small Trigger Only} & {\small Emotion Style} & {\small Emotion + Text}
\end{tabular}

\vspace{4pt}
\begin{tcolorbox}[colback=white, colframe=gray!60, boxrule=0.5pt, top=3pt, bottom=3pt, left=3pt, right=3pt]
\small
\textbf{Diary 3 (Happiness):} \begin{CJK}{UTF8}{mj}오늘 길을 걷다가 모르는 강아지가 꼬리를 흔들었다. 별거 아닌데 하루가 환해졌다.\end{CJK}\\
\textit{A stranger's dog wagged its tail at me on the street. It was nothing, really, but it brightened my whole day.}
\end{tcolorbox}

\vspace{2pt}
\begin{tabular}{ccc}
\includegraphics[width=0.3\columnwidth]{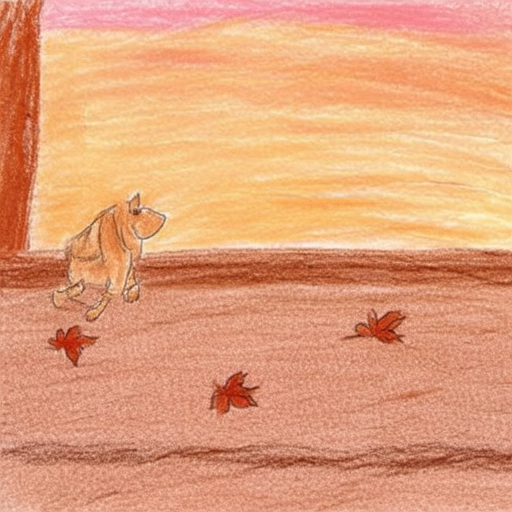} &
\includegraphics[width=0.3\columnwidth]{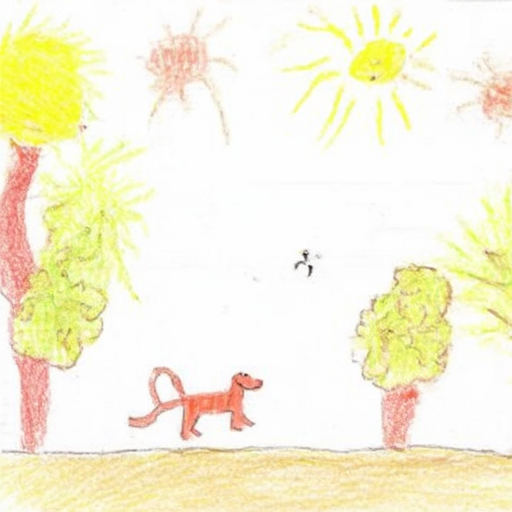} &
\includegraphics[width=0.3\columnwidth]{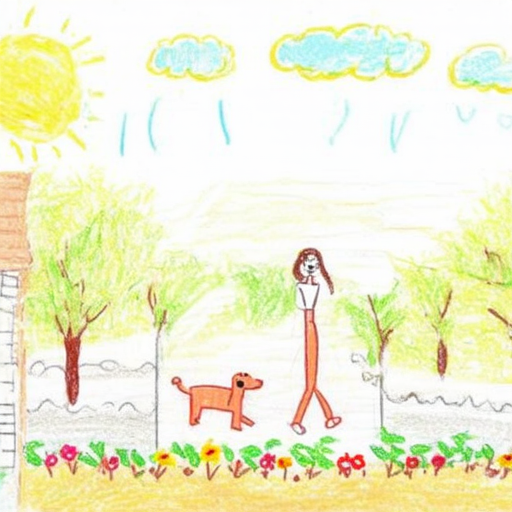} \\
{\small Trigger Only} & {\small Emotion Style} & {\small Emotion + Text}
\end{tabular}

\caption{Qualitative comparison of generated images across three LoRA caption strategies.}
\label{fig:qualitative}
\end{figure}

Although the Trigger Only strategy achieved the highest CLIP Score, the Emotion Style and Emotion + Text strategies produced images with clearer and more explicit emotional expression. As shown in Fig.~\ref{fig:qualitative}, for the second diary entry describing a closed shop (a negative experience), the Trigger Only strategy generated a warm-toned storefront with little emotional cue, whereas the Emotion Style and Emotion + Text strategies produced darker, more melancholic scenes reflecting the sadness trigger. In contrast, for the third diary entry describing a dog wagging its tail (a positive experience), the Emotion Style and Emotion + Text strategies rendered brighter and more cheerful compositions compared to the Trigger Only strategy. These results suggest that CLIP Score alone is insufficient for evaluating emotional expression in style-transferred images, as it primarily measures text-image alignment rather than the quality of style transfer or emotional representation.

\subsection{Prompt Language Experiment}

\begin{figure}[htbp]
\centering
% Replace with actual image filenames
\begin{tabular}{ccc}
\includegraphics[width=0.3\columnwidth]{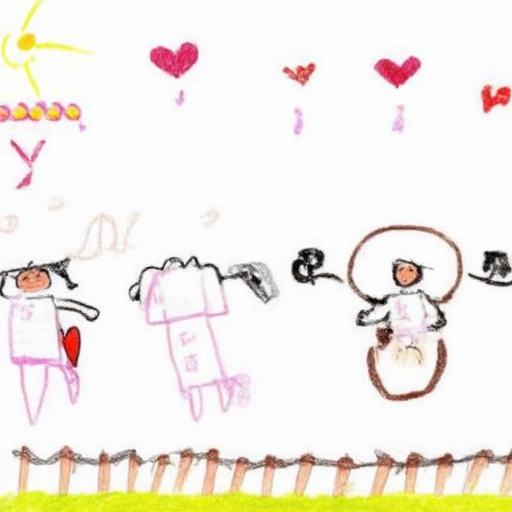} &
\includegraphics[width=0.3\columnwidth]{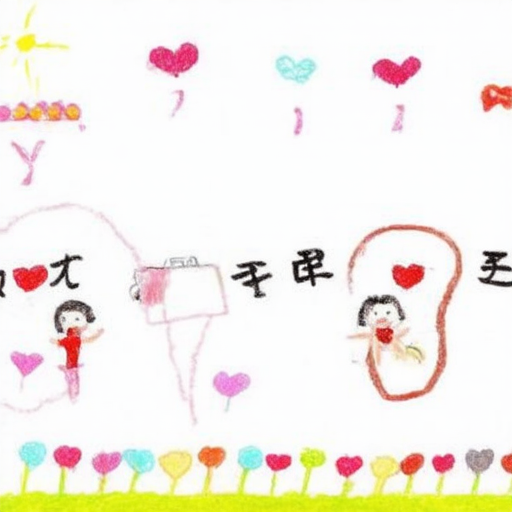} &
\includegraphics[width=0.3\columnwidth]{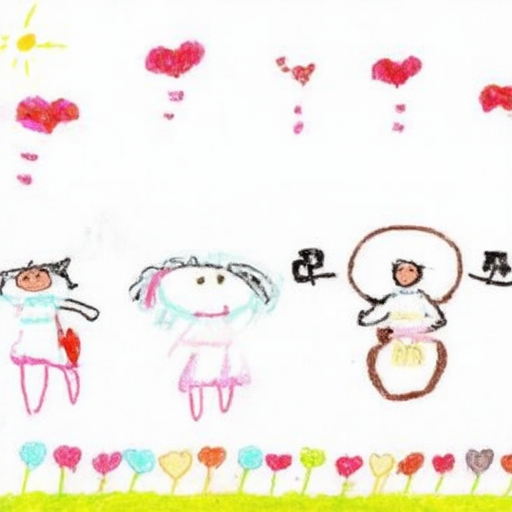} \\
\end{tabular}
\caption{Images generated from raw Korean diary text without English translation.}
\label{fig:korean}
\end{figure}

To examine the effect of prompt language, an additional experiment was conducted using raw Korean diary text as the input prompt without English translation. The generated images showed repetitive and semantically meaningless results, as SD 3.5 Medium cannot process Korean text. This confirms that LLM-based English prompt translation is an essential component of the proposed pipeline. Fig.~\ref{fig:korean} shows examples of images generated from raw Korean prompts.

% -------------------------------------------------------
\section{Conclusion}

This paper proposed an emotion-aware pipeline for generating children's hand drawing style images from short Korean diary entries. The pipeline consists of two stages: sentiment recognition and English prompt generation using Qwen3-8B, and emotion-aware image generation using Stable Diffusion 3.5 Medium fine-tuned with LoRA on a children's drawing dataset.

Through experiments, it was demonstrated that emotion-based trigger words significantly affect the visual emotional expression of generated images, and that the Emotion Style and Emotion + Text strategies produced more emotionally expressive results compared to the Trigger Only strategy. Additionally, it was found that CLIP Score alone is insufficient for evaluating style transfer quality, as it primarily measures text-image alignment rather than emotional representation. Furthermore, the prompt language experiment confirmed that English prompt translation is an essential component of the pipeline, as SD 3.5 Medium cannot process Korean text directly.

Future work will explore direct Korean language recognition and the accurate rendering of Korean text within generated images, aiming to remove the dependency on LLM-based translation. Additionally, expanding the range of emotion categories beyond happiness and sadness to include more nuanced emotional states will be investigated.

% -------------------------------------------------------
\section*{Acknowledgment}

This work was supported by the Institute of Information \& Communications Technology Planning \& Evaluation (IITP)--Innovative Human Resource Development for Local Intellectualization program grant funded by the Korea government (MSIT) (IITP-2026-RS-2022-00156334).

% -------------------------------------------------------

\end{document}